# Investigating Manifold Neighborhood size for Nonlinear Analysis of LIBS Amino Acid Spectra

Piyush K. Sharma[1], Gary Holness[2], and Poopalasingam Sivakumar, Yuri Markushin, NoureddineMelikechi[3]

[1]Department of Mathematical Sciences, Delaware State University, USA
[2]Department of Computer and Information Sciences, Delaware State University, USA
[3]Department of Physics and Engineering, Delaware State University, USA,
pksharma12@students.desu.edu,(gholness, psivakumar, ymarkushin, nmelikechi)@desu.edu

**Abstract**

Classification and identification of amino acids in aqueous solutions is important in the study of biomacromolecules. Laser Induced Breakdown Spectroscopy (LIBS) uses high energy laser-pulses for ablation of chemical compounds whose radiated spectra are captured and recorded to reveal molecular structure. Spectral peaks and noise from LIBS are impacted by experimental protocols. Current methods for LIBS spectral analysis achieves promising results using PCA, a linear method. It is well-known that the underlying physical processes behind LIBS are highly nonlinear. Our work set out to understand the impact of LIBS spectra on suitable neighborhood size over which to consider pattern phenomena, if nonlinear methods capture pattern phenomena with increased efficacy, and how they improve classification and identification of compounds. We analyzed four amino acids, polysaccharide, and a control group, water. We developed an information theoretic method for measurement of LIBS energy spectra, implemented manifold methods for nonlinear dimensionality reduction, and found while clustering results were not statistically significantly different, nonlinear methods lead to increased classification accuracy. Moreover, our approach uncovered the contribution of micro-wells (experimental protocol) in LIBS spectra. To the best of our knowledge, ours is the first application of Manifold methods to LIBS amino-acid analysis in the research literature.

## 1 Introduction

LIBS is a powerful analytical technique that provides information about the elemental composition of a given sample [1], [2]. The method uses an intense short laser pulse to break down the matrix of the target and to create a short-lived micro-plasma. During the cooling of plasma, atomic, ionic, and occasionally molecular constituents emit the discrete spectra which is further collected and analyzed, allowing characterization of the sample content Fig. 1.

LIBS applications has grown beyond elemental analysis of rocks, metals and plastics to analysis of more complex biological materials or clinical specimens [2]-[5]. Living organisms are comprised of biomacromolecules (BMs) such as proteins, polysaccharides, nucleic acids and, lipids. LIBS can be considered as a new emerging technology for in-situ detection and identification of those BMs [2], [6].

Identification of a protein within a mixture of proteins is difficult as they appear very similar when measured by emissions of their principal constituents, namely Hydrogen, Carbon, Nitrogen, and Oxygen. LIBS has achieved high detection sensitivity and classification accuracy (above 94% for 4 specific proteins detected separately) [6] - [10].

In LIBS applications, higher precision spectroscopic instruments are capable of providing large volumes of data that capture the dynamics of complex phenomena [11]. Traditional data analysis tools and the state-of-the-art techniques for analysis of spectroscopic data are based on standard linear statistical and machine learning algorithms available in commercial software packages used in spectroscopy. However, such data are difficult to analyze due to high dimensionality (number of spectral components), nonlinear dependence between relative proportions of compounds in a mixture and the resulting spectral signature, heteroscedasticity, and non-linear dynamics concerning the linkage between micro-macro scale phenomena observed in spectra of macro-molecules. This necessitates the need for non- linear machine learning and data mining techniques for large and heterogeneous spectroscopic data. This paper addresses such techniques for the analysis of spectroscopic data obtained from amino acid solutions. To the best of our knowledge, ours is the first application of Manifold methods to LIBS amino-acid analysis in the research literature.

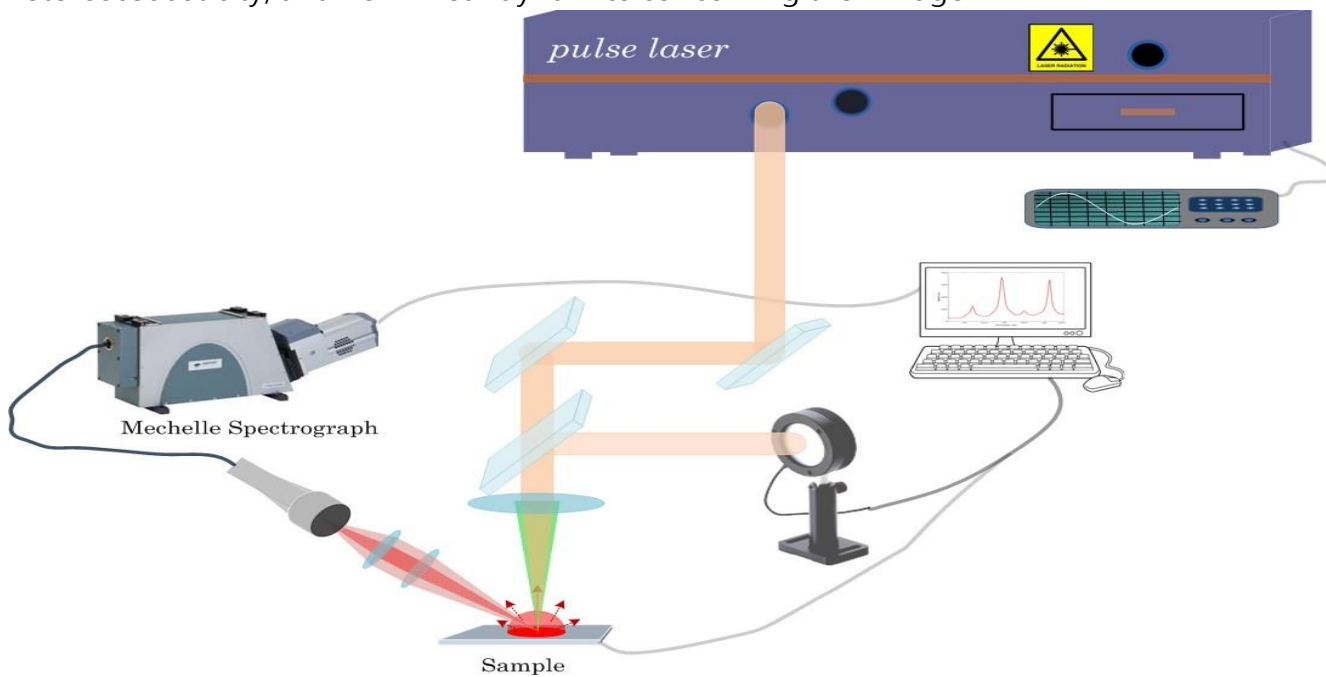

Figure 1: LIBS Experimental Setup.

This work also benefits LIBS analysis of polysaccharides, a key step in the study of the polymerization of polymeric proteins and the polymeric structure of polysaccharides and nucleic acids.

In this work we study frozen samples of Water, Polysaccharide (Ficoll), and four nonessential amino acids: Aspartic Acid (ASP), Glutamic Acid (Glu), Cysteine (Cys), and D-Serine (Ser D).

Our data consisted of a total 670 instances, each representing a single LIBS spectrum,

produced from 13 different samples drawn from the 6 aforementioned compounds. Captured wavelengths remain consistent across all LIBS spectra. This resulted in 26100 different wavelengths. In our experiment, we used the spectral intensity at each wavelength as a feature. Therefore, our data instances were vector valued in 26100 dimensions.

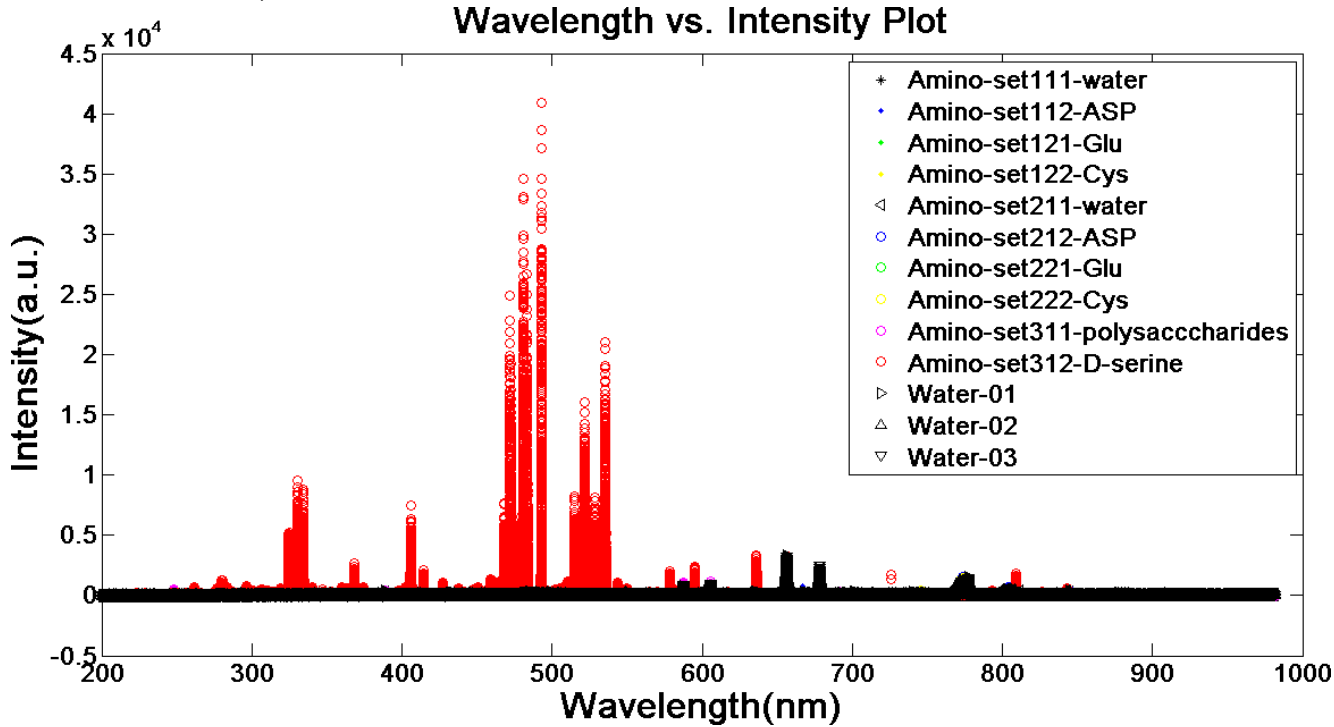

Figure 2: High intensity peaks of D-Serine (red) at certain $\lambda$ regions most likely due to micro-well influence.

## 2  Methodology

We examine the relationship between spectral emission line intensity and wavelength ($\lambda$). After partitioning the spectra into 8 equal sized regions (or bins), we compute the total spectral intensity of each region and plot their histograms with 10 bins Fig. 3. Treating each histogram as a distribution, we computed its expected intensity (E.I.) Tab. 1.

We employ PCA & cMDS for linear dimensionality reduction, and ISOMAP & LLE for nonlinear dimensionality reduction [12] - [19]. We visualize the reduced version of our data as well as employ quantitative methods for analysis. To visualize, we produce 3D plots using the principle components for the first 3 dimensions.  Quantitatively, we generate scree plot (residual variance) and Davies-Bouldin Criterion/Index (DBI) plots [20]. DBI was parameterized using k-means clustering for 10 clusters. We developed a more fine grained spectrum analysis method using Shannon Entropy over all 26100 wavelengths to compute a density measure, something we call Entropy Density [21].

Finally, we perform SVM classification using Weka's SMO optimization algorithm and validate through 10 fold cross-validation and polynomial kernels of degree K=1:5. Classification accuracies are reported for the unprocessed dataset $X$, and for the dimension reduced (d=1:10) versions of the datasets. For classification results, statistical significance is indicated through errorbars representing 1 standard deviation.

We used the cMDS, ISOMAP, and LLE dimensionality reduction implementations from the EDA Toolbox (http://pi-sigma.info) along with our own MATLAB implementations [22].

## 3 Discussion of Analysis

### 3.1 Wavelength-Intensity Linkage

Water and Polysaccharide are the only compounds that had $-ve$ values of the total intensity within the spectral region $199-299 nm$. Similar behavior corresponding to similar E.I. in the majority of compounds except D-Serine whose intensity is higher than other compounds in many spectral regions (199–599.02). All compounds have the similar E.I. intensities in region 899.16 981.54. The significant spectral intensity peaks of D-Serine samples are consistent with brass alloy (Cu, Zn) micro-well sample holder material. The detection of spectral lines Fig. 2 of Cu I (324.754 nm, 327.396 nm, and 521.820 nm) and Zn I (334.501 nm, 330.258 nm, and 481.053 nm) describes the micro-well chemical composition. A review of experimental protocol has identified brass alloy (Cu, Zn) micro-wells whose composition coincides with significant peaks observed in D-Serine spectra.

Table 1: Expected Intensities and Spectral Ranges.

| $\lambda$ | Water | ASP | Cys | Glu | Polysac. | Ser_D |
|---:|---|---|---|---|---|---|
| 199 - 299 | 5.84e+04 | 6.18e+04 | 5.84e+04 | 6.23e+04 | 5.11e+04 | 1.28e+05 |
| 299.01 - 399.02 | 1.40e+05 | 1.40e+05 | 1.37e+05 | 1.41e+05 | 1.25e+05 | 4.68e+05 |
| 399.04 - 499.03 | 1.56e+05 | 1.52e+05 | 1.49e+05 | 1.50e+05 | 1.44e+05 | 8.04e+05 |
| 499.06 - 599.03 | 1.38e+05 | 1.38e+05 | 1.37e+05 | 1.37e+05 | 1.36e+05 | 5.17e+05 |
| 599.06 - 699.03 | 1.75e+05 | 1.60e+05 | 1.53e+05 | 1.52e+05 | 1.33e+05 | 1.67e+05 |
| 699.08 - 799.07 | 1.68e+05 | 1.92e+05 | 1.81e+05 | 1.75e+05 | 1.23e+05 | 1.21e+05 |
| 9.12 - 899.11 | 1.18e+05 | 1.31e+05 | 1.26e+05 | 1.24e+05 | 9.83e+04 | 1.11e+05 |
| 899.16 - 981.54 | 7.47e+04 | 7.61e+04 | 7.55e+04 | 7.63e+04 | 6.78e+04 | 7.8e+04 |

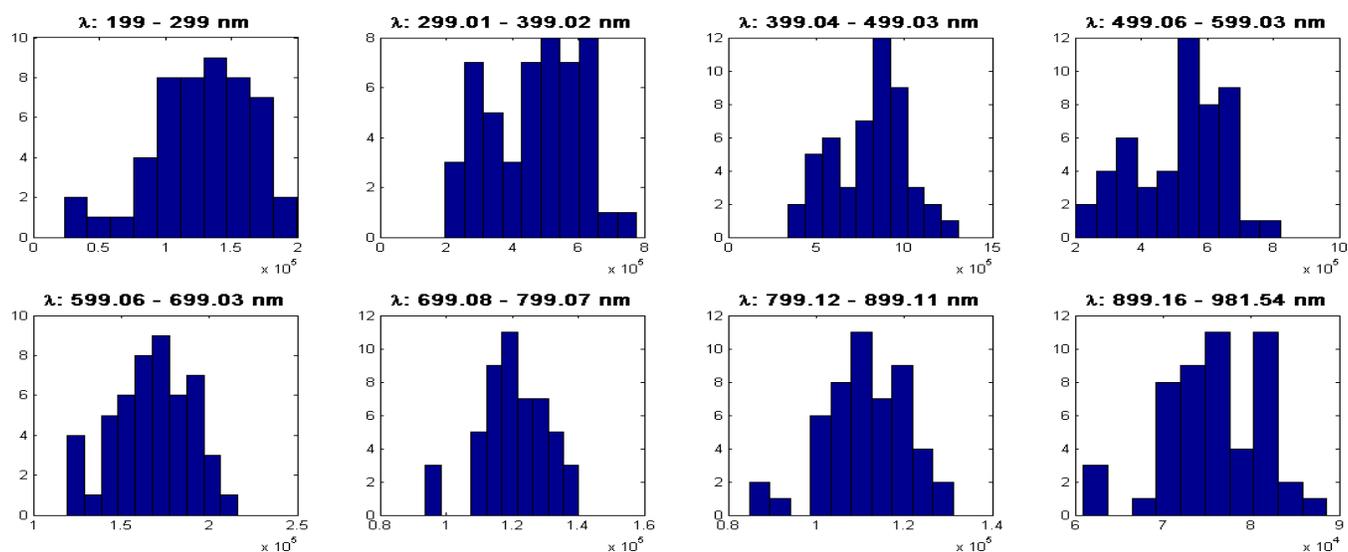

Figure 3: Histograms of total intensity (on x-axis) within specified spectral regions for D-Serine.

## 3.2 Pattern Recognition

| | |
|---|---|
| ✱ | Amino-set111-water |
| ◆ | Amino-set112-ASP |
| ◆ | Amino-set121-Glu |
| • | Amino-set122-Cys |
| ◁ | Amino-set211-water |
| ○ | Amino-set212-ASP |
| ○ | Amino-set221-Glu |
| ○ | Amino-set222-Cys |
| ○ | Amino-set311-polysacccharides |
| ○ | Amino-set312-D-serine |
| ▷ | Water-01 |
| △ | Water-02 |
| ▽ | Water-03 |

The basic problem is to reduce the size of a high dimensional data thus making it easier to analyze. However, we would also like to preserve most of the information represented by this data which may be lost during dimensionality reduction. We seek to extract only the features necessary for patterns that discriminate between compounds.

PCA and cMDS are linear methods that employ pairwise comparisons. PCA uses covariance (statistical linkage) while cMDS uses pairwise distances or dissimilarities. As linear methods, PCA & cMDS both use Euclidean distance, and therefore, yield similar results. ISOMAP uses pairwise comparisons but within a neighborhood along the manifold. LLE uses set of neighbors that are the k-closest, but represents manifold structure using weight matrix encoding the reconstruction of each point by linear combination of its neighbors. We use linear methods as the benchmark. These encode global relationships (between all pairs of data points whether near or far).

We use ISOMAP because it employs comparisons among data points covering increasingly large patches of manifold surface (i.e. Larger neighborhood). This allows us to examine how neighborhood size, ranging from local to global, is able to capture pattern phenomena. As LIBS is nonlinear, the question we enterprise to address concerns what is a suitable resolution or scale at which to capture pattern variations in LIBS spectra. ISOMAP captures pairwise distance between data points within a neighborhood, but ignores extra information encoded in relative relationship within constellations of data points. Therefore, we use LLE because it employs neighborhoods but measures the relative relationship among data points.

In our experiments, we consider relationships in increasingly large neighborhoods. This makes use of additional information encoding constellations of points in a patch on the manifold versus pairwise relationships between points in a patch. One issue with nonlinear methods is to define a neighborhood size which is selected based on intuition and experimentation. A poor choice leads to a disconnected graph approximation yielding bad results. One of the main goals of this paper is the investigation of more informed choice in neighborhood size.

D-Serine's variability is the most distinguishable among the compounds. This held across all dimensionality reduction models and entropy density plots. Water and Polysaccharide were also easily identified. However, ASP, Glu and Cys are highly overlapping, thus, hard to discriminate Fig. 5, Fig. 6. Color labels used in 3D scatter plots are listed in Fig. 4.

Figure 4: Color Labels.

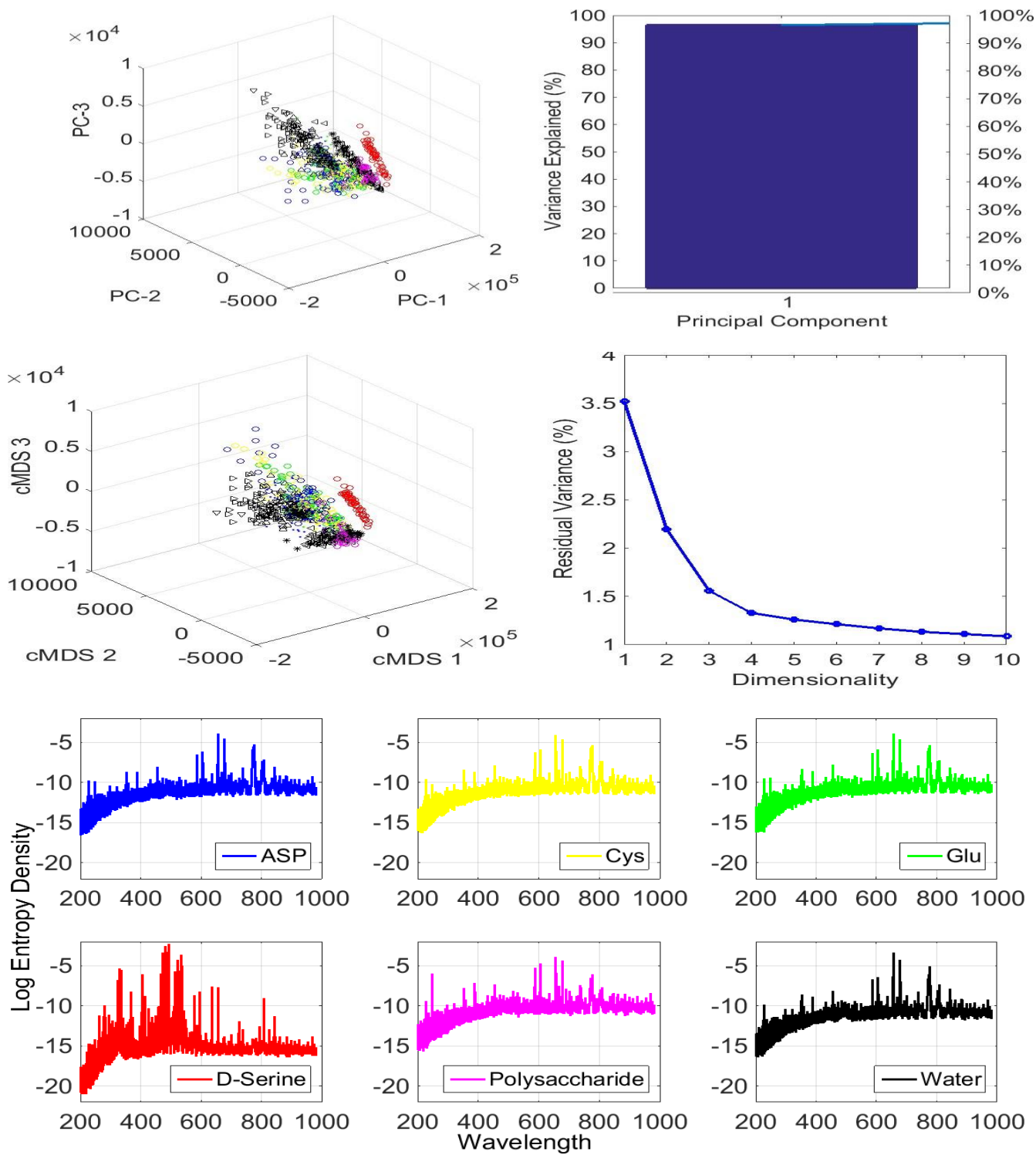

Figure 5: PCA (Top) & cMDS (Middle) 3D Scatter and Scree Plots. D-Serine (red) can be seen varying separately from the other compounds. PCA & cMDS explained 96.47% & 96.5% of the variance respectively. Notice a thin blue line corresponding to the 1st PC, and the elbow of the curve at dimensionality 3. Wavelength vs. log Entropy Density Plots (Bottom) for each amino acid dataset. Notice highest peaks appear for D-Serine (red) ranging up to
−2.5. Best seen in colors.

Table 2: Explained Variance Summary.

| Model | Variance Explained (%) | Principal Components |
|---|---|---|
| PCA | **96.47%** | 1 |
| cMDS | **96.5%** | 1 |
| ISOMAP | **98.8% - 99.93%** | 1 |

## 3.3 Clustering

Davies-Boulding Index (DBI) is defined as a function of the ratio of the within cluster scatter to the between cluster separation. Therefore, a lower value of DBI implies the better clustering. This occurs when clusters are compact and well separated. A good dimensionality reduction technique should preserve most of the significant information and find low dimensional embedding with similar characteristics like in the original dataset. We use this concept to cluster the reduced datasets obtained by different methods. A better method should give more distinct clusters and low DBI. We use this approach in particular to compare the descriptive power of LLE with other methods which provide residual variances while LLE does not.

Table 3: Only 5 dimension of LLE embedding with neighborhood size $k = 30$ were sufficient for k-Means to find 6 Clusters for six amino-acid classes (See the Corresponding Minimum Davies-Bouldin Index in the third column of the last row of Fig. 7).

| Model | DBI | Dimensions | Clusters | Neighborhood |
|---|---|---|---|---|
| LLE | 0.52 | 5 | 6 | 30 |

As one of main goals of this paper, we vary neighborhood size in order to understand the impact of LIBS spectra on suitable manifold $k$-neighborhood size. The use of the DBI score on a suitable neighborhood size, evaluates the methods' ability to tease apart inter-sample variability from intra-sample variability.

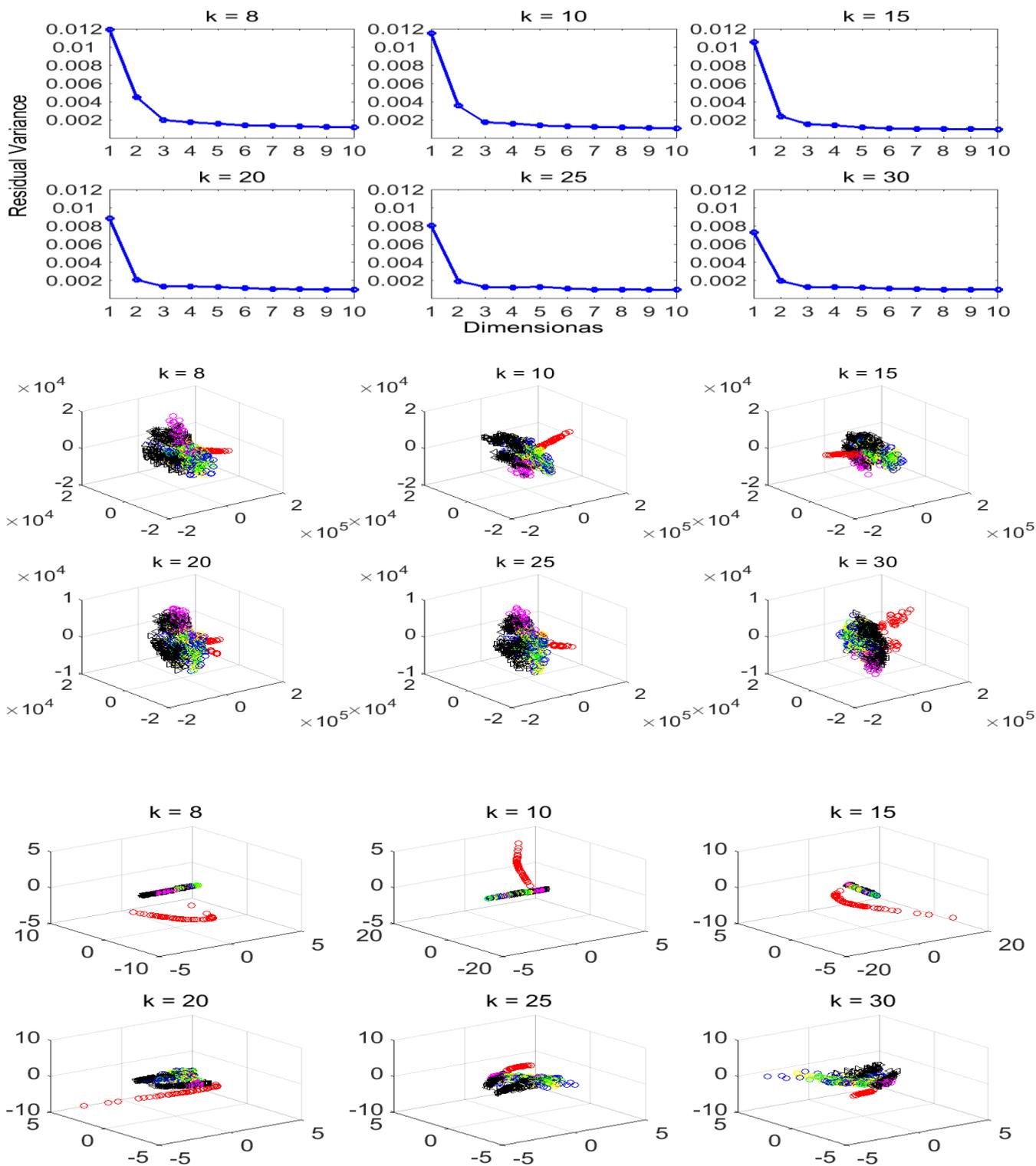

Figure 6: ISOMAP Scree Plots (Top). Notice a decline in residual variance with $k$ increasing, and the elbow of the curve unaltered at dimension $d = 3$. ISOMAP (Middle) & LLE (Bottom) 3D Scatter Plots. Notice D-Serine's variation from one direction to other for $k > 15$.

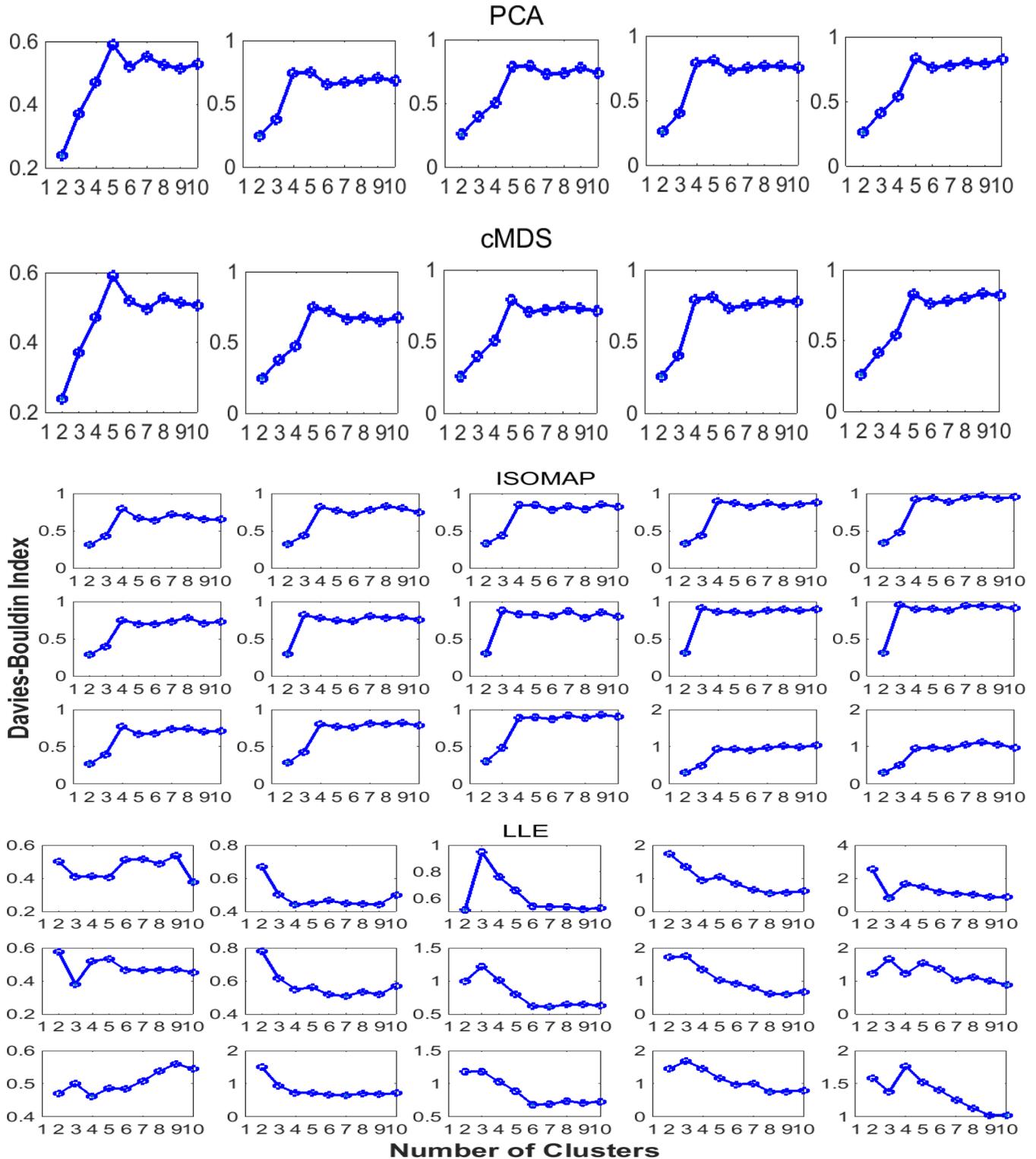

Figure 7: Clusters vs. DBI plots for each dataset. From left to right, plots correspond to 2, 3, 5, 7, & 10 dimensionsrespectively. The 1st, 2nd, & 3rd rows of ISOMAP & LLE plots correspond to $k$ = 8, 15, & 30.

## 3.4 Classification

Finally, to test our assertion that pattern phenomena in LIBS are nonlinear, we perform classification on embedded dimensions $d$ using SVMs with increasing polynomial kernel size $K$. A polynomial of higher order should give better classification accuracy in contrast to a polynomial of lower order. Moreover, as one of our goals is to understand the impact of LIBS spectra on a suitable neighborhood size, we select different manifold neighborhood sizes for ISOMAP& LLE.

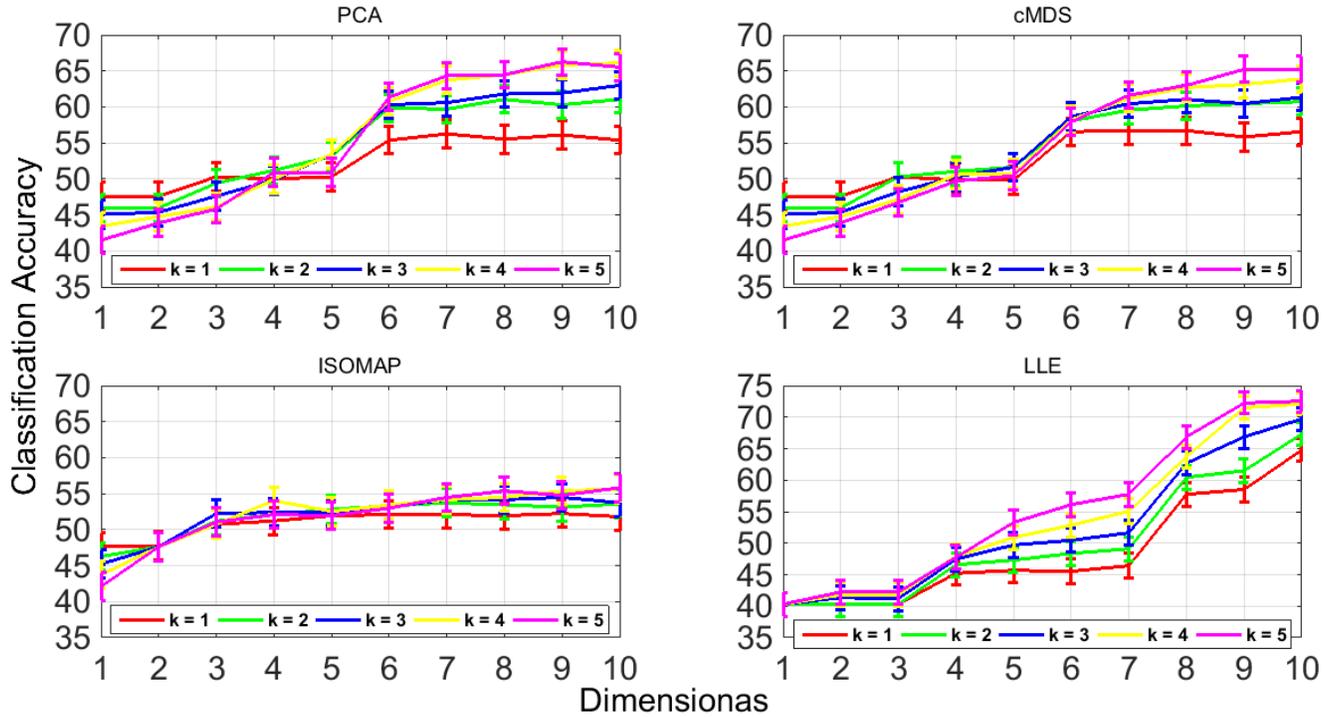

Figure 8: Errorbar Plots for each model. Here manifold nbd size is 200. Notice accuracy increasing with $d$ & $K$
increasing (highest for $K$ = 5 (magenta)) suggesting the intrinsic nonlinearity associated to LIBS dataset.

Table 4: Highest Classification Accuracy Summary.

| Model | Accuracy(%) | Kernel | Dimensions | Std-Dev | nbd |
|---|---|---|---|---|---|
| PCA | **66.2687%** | 5 | 9 | **1.827%** | N/A |
| cMDS | **66.2687%** | 5 | 9, 10 | **1.840%** | N/A |
| ISOMAP | 63.8806% | 5 | 8 | 1.856% | 10 |
|  | **64.1791%** | 4 | 9 | **1.930%** | 20 |
|  | 62.8358% | 5 | 10 | 1.913% | 30 |
|  | 56.2687% | 5 | 10 | 1.910% | 100 |
| | 55.8209% | 5 | 10 | 1.907% | 200 |
| | 66.7164% | 5 | 10 | 1.895% | 10 |
| | 63.4328% | 4 | 10 | 1.894% | 20 |
| LLE | 62.2388% | 5 | 9 | 1.897% | 30 |
| | 69.8507% | 5 | 10 | 1.897% | 100 |
| | **72.5373%** | 5 | 10 | **1.895%** | 200 |

# 4 Conclusion

One of our experimental goals was to explore the impact of spectral energy on the requisite neighborhood size for which to uncover pattern phenomena. Confirming our assertion, experiments demonstrated differences in energy distribution in LIBS spectra impacted neighborhood size for manifold methods we employed. In order to compute the connected graph required as a prerequisite for manifold construction, a minimum $k$-neighborhood size of 8 was required. As the neighborhood size increases, manifold methods increasingly capture global smoothness trends. This points to our assertion that pattern phenomena are non-linear as LIBS spectra require relatively larger neighborhoods in order to capture large enough patches of manifold surface to draw distinctions. We also found that, in the case of ISOMAP and LLE, beyond a $k$-neighborhood size of 15, discernible intra-sample type (same compound) differences arose. Likewise for $k$ below 15, discernible inter-sample (different compound) differences arose. Therefore, in using manifold methods on such data, a good choice of $k$-neighborhood size should capture both inter-sample and intra- sample differences. The neighborhood parameter would fall in the interval [8, 15]. An ideal method, for future research, would employ multiple neighborhood sizes.

By considering expectation of intensities for grouped spectral regions, we discriminated between D-Serine and the other compounds. Moreover, the contribution of brass alloy (Cu, Zn) micro-wells to spectral intensity was easily observable. Our approach of using expectation of intensity not only discriminates compounds, but also confirms our results on the effect of differences in spectral intensity distributions on suitable manifold neighborhood size. Energy is absorbed and radiated back as a function of atomic orbital structure Tab. 1. It is precisely the differences in such distributions that describe and explain a compound's chemical structure.

We found that PCA explained 96.47%, and cMDS explained 96.5% of the variance using the 1st principal component. While this representation is compact, ISOMAP explains 98.8% ($k$ = 8) through 99.93% ($k$ = 30) of the variance by a single component Tab. 2.

For visualization, Scatter plots and log Entropy Density clearly discriminated D-Serine from the other compounds. PCA, cMDS, and ISOMAP visualizations provide clear discrimination between amino acids, Polysaccharide and Water Fig. 5, Fig. 6.

DBI and residual variance quantitatively verify observations in the visualizations. DBI confirms that LLE supports the formation of clusters that have smaller intra-cluster variability and larger inter-cluster variability (more compact clusters spaced further apart). Among all variations of dimensionality for each embedding, only 5 dimension of LLE with neighborhood size $k = 30$ were sufficient for k-Means to find 6 clusters for six amino-acid classes Tab. 3. The corresponding DBI (minimum is better) can be seen the third column of the last row of Fig. 7).

Classification results suggest that higher dimensionality yields better performance. Support Vector Machine (SVM) employ kernels for projection of data into nonlinear feature spaces. From our experiments, SVM employing polynomial kernels of increasing degree resulted in better classification accuracy. Among tested methods LLE resulted in the best classification accuracy for SVM; attributable to LLE capturing additional information on relative relationships among constellations of points. This points to **our assertion that LIBS pattern phenomena are nonlinear** Fig. 8, Tab. 4.

Lastly, all our methods confirmed a different spectral response for D-Serine versus the other tested compounds. Further investigation confirmed a difference in sample preparation of D-Serine. This experimental protocol employed micro-wells. This demonstrated the utility of our methods as they identified a change in experimental protocol across tested compounds. While the impetus behind this work began with LIBS analysis of amino-acid spectra, we believe these findings are applicable beyond amino-acid analysis.

We will extend these methods to the characterization of the remaining non-essential amino acids (alanine, arginine, asparagine, glutamine, glycine, proline, tyrosine). This will build towards the goal of finding relationships between amino acid building blocks & polypeptides as revealed through their LIBS spectra, something we call micro-macro scale phenomena. Further, we will investigate new methods for multiresolution analysis on manifolds. In doing so, we enterprise to develop methods effective in capturing both inter & intra sample variability.